# Embedding Large Language Models into Flow Controls: An Agentic Framework for Adaptive and Trustworthy Automated Cooking

Zihan Song, Hongwei Huang, Yueshuo Sun, Yonglin Tian, Fei-Yue Wang, and Bai Li

***Abstract*—Automated cooking robots have traditionally relied on predefined procedures and rule-based control, ensuring stable execution but offering limited personalization, whereas recent large-model approaches support natural language interaction but often suffer from opaque decision making and unreliable execution in real kitchens. To address this challenge, this paper proposes an agentic framework that systematically decomposes personalized cooking requirements into structured and verifiable control programs rather than directly mapping language to actions. Multiple AI agents collaboratively transform user intents into canonical recipes, workflow programs with explicit flow control, and executable Python code grounded in an atomic action library. The system consists of three tightly coupled stages: offline recipe-to-code generation through multiple agents, online closed-loop execution with supervisory intervention enabled by multimodal perception, and post-run adaptation that updates user preference models for long-term personalization. Real-world experiments on a physical cooking platform demonstrate that the proposed framework achieves reliable task completion, transparent execution logic, and effective anomaly handling across diverse personalized scenarios, validating its practicality for trustworthy automated cooking in real environments.**

## I. Introduction

COOKING has long relied on human labor, yet human-centered cooking suffers from two inherent limitations. First, cooking outcomes are difficult to reproduce consistently: different chefs naturally produce different results, and even the same chef may vary across time or conditions, leading to unstable taste and quality. Second, human cooking entails high labor and training costs, frequent turnover, and limited process control, while subjective judgment and convenience-driven decisions further introduce uncertainty [1]. As industrial automation advances, it is increasingly inevitable that machines assume cooking tasks requiring consistency, repeatability, and cost efficiency [2], much like the ongoing transition from manual to autonomous driving.

In response, automated cooking robots have gradually emerged. Early systems primarily focused on basic automation through demonstration, rule-based control, or predefined workflows, enabling stable dish reproduction and partial cost reduction. However, such systems typically produce uniform outputs and cannot accommodate individual preferences related to taste, dietary habits, or health constraints. As expectations evolve, automated cooking systems are required not only to operate reliably but also to adapt flexibly to personalized demands [3].

Recent advances have introduced large language models (LLMs) and embodied intelligence into cooking automation, allowing systems to interpret natural-language requests and generate complete cooking actions end to end. While these approaches appear to support personalization, they largely function as black-box decision-makers, lacking explicit intermediate representations and auditable execution logic, which raises concerns about reliability and safety in real cooking environments [4].

To mitigate these issues, some studies have explored structured and interpretable representations of recipes, decomposing cooking processes into explicit workflows and atomic actions. Although such methods improve transparency at the planning level, they are often limited to offline execution and lack mechanisms for online sensing and feedback correction [5]. When real-time adjustments are incorporated, interpretability is frequently lost again.

Thus, the existing automated cooking solutions have yet to simultaneously achieve automation, personalization, transparent execution, and online adaptivity throughout the cooking process. Addressing these requirements within a unified and practical framework remains an open challenge.

### *A. Contributions*

Contributions of this work are summarized as follows.

1) We propose an agentic framework for automated cooking that establishes a transparent pathway from personalized natural-language requirements to executable robotic behaviors. Instead of directly generating actions in an end-to-end manner, the proposed framework progressively transforms user intents into canonical recipes, structured workflows with explicit flow-control constructs, and executable Python programs grounded in an atomic action library, enabling interpretable, verifiable, and trustworthy cooking automation.

2) We develop an integrated closed-loop cooking system that combines AI-generated programs, multimodal perception,

This work was supported by National Natural Science Foundation of China under Grant 62103139, Lotus Youth Talent Program of Hunan Province, China under Grant 2023RC3115, and Science and Technology Commission of Shanghai Municipality under Grant 22DZ2229004.

Zihan Song and Yueshuo Sun are with the College of Mechanical and Vehicle Engineering, Hunan University, Changsha 410082, China (e-mails: songzihan@hnu.edu.cn, yssun@hnu.edu.cn).

Hongwei Huang and Bai Li are with the School of Information and Electronic Engineering, East China Normal University, Shanghai 200241, China (e-mails: 51285904023@stu.ecnu.edu.cn, libai@zju.edu.cn).

Yonglin Tian and Fei-Yue Wang are with the State Key Laboratory for Management and Control of Complex Systems, Institute of Automation, Chinese Academy of Sciences, Beijing 100190, China (e-mails: yonglin.tian @ia.ac.cn, feiyue.wang@ia.ac.cn).

asynchronous supervisory intervention, anomaly recovery, recipe migration, and post-cooking preference adaptation. By tightly coupling planning, execution, and personalization within a unified architecture, the proposed system achieves reliable task execution and adaptive personalized cooking in real-world environments.

### *B. Organization*

The remainder of this paper is organized as follows. Section II reviews representative approaches to automated cooking and robotic chef systems. Section III presents the overall architecture of the proposed automated cooking system. Sections IV to VI describe the three modules of the proposed framework, including pre-cooking preparation, online execution, and post-cooking adaptation. Section VII introduces the construction of the cooking knowledge dataset and the LoRA-based supervised fine-tuning procedure. Section VIII presents the experimental results and discussions. Finally, Section IX concludes the paper.

## II. Related Work

This section reviews representative automated cooking approaches in the literature, which are broadly classified into three categories according to their control paradigms and levels of intelligence.

### *A. Automated Cooking via Rule-/Script-Based Controls*

Early cooking systems focused on automated execution of predefined procedures using rule-based or script-driven control, with an emphasis on consistency and repeatability.

Mondal et al. [6] implemented one of the most direct realizations of this paradigm by encoding complete dishes as fixed sequential programs executed by microcontrollers. After a user selected a menu item, the robot executed a hard-coded sequence of operations, including dispensing, heating, stirring, and shutdown. Yan et al. [7] followed a similar script-driven approach but enhanced execution precision through teach-and-playback motion primitives, quantitatively controlled ingredient dispensing, and online temperature monitoring, thereby improving repeatability while maintaining an offline, procedure-centric control structure.

To overcome the limitations of flat action scripts, Ma et al. [8] modeled cooking processes using hierarchical task networks (HTNs), representing each recipe as a layered task structure that decomposed fixed dishes into semantically organized subtasks and atomic actions. Building on structured task representations, Noh et al. [9] introduced gate conditions and resource-aware scheduling to coordinate multi-arm manipulation and appliance usage, formulating multi-dish preparation as a constrained scheduling problem under fixed environmental assumptions. Sugiura et al. [10] extended scripted execution to household environments by compiling user-specified timelines for ingredient addition and heating into control scripts that coordinated mobile robots, while keeping the procedural structure unchanged during execution.

Several studies further enriched script-based cooking by refining execution quality without modifying task structure. Junge et al. [11] parameterized key variables within fixed omelette-cooking scripts, enabling optimization over seasoning and timing while preserving the predefined action sequence. Sochacki et al. [12] extracted atomic actions from human demonstrations using perception and probabilistic filtering, yet ultimately executed cooking through predefined programmatic modules. Sochacki et al. [13] and Shi et al. [14] integrated taste-related sensory feedback into scripted procedures, adjusting parameters such as salt or water addition via controllers, while retaining an open-loop action structure.

The aforementioned studies demonstrated that rule- and script-based approaches can achieved automated cooking with high consistency and robustness. However, recipes, task structures, and execution logic were largely fixed *a priori*, limiting adaptability to user preferences, contextual variations, and dynamic cooking requirements.

### *B. Automated Cooking with End-to-End Embodied Intelligence*

Recent advances in large models and embodied intelligence have enabled end-to-end frameworks that directly map natural language instructions to cooking actions, thereby supporting personalization and task generalization.

Driess et al. [15] proposed PaLM-E, which unified language, vision, state, and control within a single model, enabling direct inference from user intent to continuous robot actions without relying on explicit rules or programs. Building on this paradigm, Brohan et al. [16] introduced RT-2 by augmenting vision–language models with action tokens, allowing multimodal inputs to be translated directly into robot commands through large-scale cross-task training, including cooking-related manipulation tasks.

These end-to-end methods remain largely black-box, providing limited transparency and fine-grained control over cooking processes, which poses challenges for reliable online supervision and practical deployment.

### *C. Automated Cooking with AI-Guided Structured Programs*

This subsection reviews AI-guided structured-program approaches that improve transparency and controllability of automated cooking by introducing explicit intermediate representations, such as symbolic task models, executable programs, and logic-based constraints, thereby making system behavior interpretable and inspectable during execution.

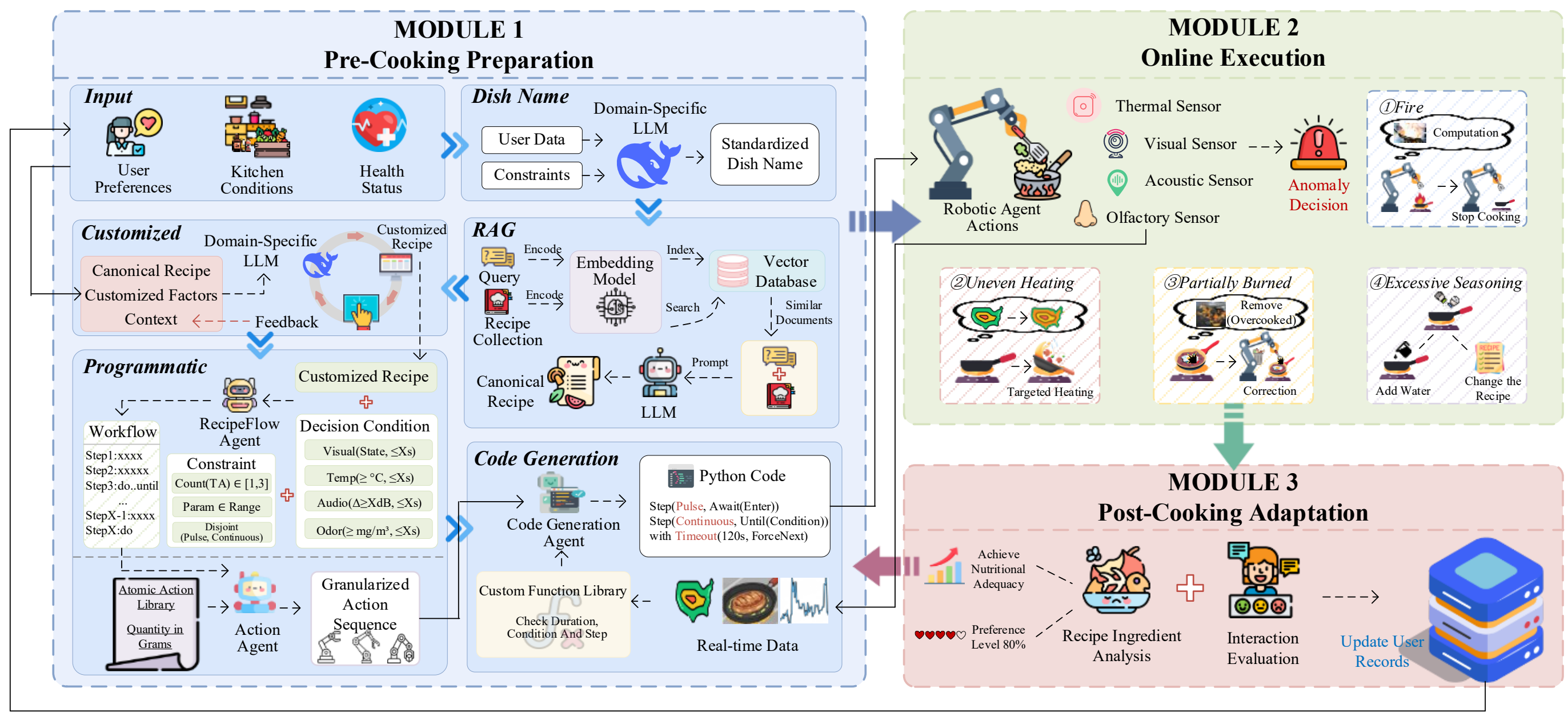

Fig. 1. Overall architecture of the proposed automated cooking system.

Gravot et al. [17] formulated cooking as a hierarchical task network planning and execution framework, in which partially ordered recipe tasks were decomposed into symbolic primitives bound to callable control procedures, with a supervisory layer monitoring execution and triggering replanning upon symbolic state changes or failures; however, both recipes and decomposition rules were largely hand-specified, personalization was limited, and adaptation focused on symbolic consistency rather than continuous process feedback. Schmitz et al. [18] transformed motion-captured human demonstrations into explicit PDDL domains by classifying predefined side actions and clustering main processing actions, enabling procedural planning over structured symbolic representations; however, the approach relied on constrained environments and offline processing, with state semantics and updates restricted to predefined categories. Inagawa et al. [19] translated Japanese text recipes into pseudo-executable code through morphological analysis and mappings to predefined motion libraries, followed by offline motion planning; however, fixed recipe formats and environments limited perception-driven adaptation in cooking.

Vemprala et al. [20] leveraged LLMs to generate structured outputs, such as code snippets and API-level function calls, that explicitly orchestrated perception, planning, and control modules, keeping task logic readable and auditable; however, execution depended mainly on offline reasoning and dialog-based correction, leaving online sensory feedback loosely coupled to program execution. Liang et al. [21] proposed Code as Policies, in which language models generated executable Python programs with explicit control flow that queried perception APIs and invoked parameterized motion primitives; however, robustness was bounded by the completeness of available primitives, and readable code alone did not guarantee reliable closed-loop process regulation. Yoneda et al. [22] further externalized decision making through executable code paired with an explicitly maintained world state, using separate components to generate actions and update symbolic state from observations; however, maintaining faithful state estimates for continuous cooking variables such as doneness or mixing uniformity remained challenging.

Mavrogiannis et al. [23] compiled online recipe instructions into auditable intermediate programs using slot-based semantic parsing and temporal logic constraints, enabling structured decomposition into admissible primitives with reuse through caching; however, decomposition errors could propagate through execution, and semantic parsing required manual annotation with limited coverage of cooking behaviors. Verghese and Atkeson [24] represented cooking skills as selecting interpretable behavior templates from a library of parameterized hybrid position and force controllers guided by online information sources, preserving execution structure; however, dependence on fixed templates and stable execution conditions limited adaptability, and template selection often required trial execution, reducing efficiency and robustness.

## III. Overall Automated Cooking Architecture

This section presents our automated cooking framework with three modules, covering pre-cooking preparation, online execution, and post-cooking adaptation, as shown in Fig. 1.

The pre-cooking preparation module takes user information and the current cooking context as inputs and produces executable cooking code as its output. It explicitly models user intent and environmental conditions by collecting multimodal user information, including dietary preferences, health conditions, cultural background, special occasions, and historical dining records, together with the availability of ingredients, seasonings, robotic cooking capabilities, and sensor configurations. Based on these inputs, a target dish is selected and a customized recipe is generated by adapting a canonical recipe to user-specific constraints. The customized recipe is then transformed into a programmatic recipe with a prescribed structural format, where cooking steps are expressed using explicit flow-control constructs such as conditional branches and do-until patterns. Each step in the programmatic recipe is grounded in an atomic action library supported by the cooking robot and sensors, and the entire pipeline ultimately generates executable Python control code that can be compiled and run in the online execution module.

The online execution module carries out the cooking process in real time by executing the generated Python control code under continuous sensory feedback. A cooking agent performs the prescribed atomic actions through its manipulators while multimodal sensors provide synchronized observations of the cooking state, including visual, thermal, auditory, and force-related signals. In parallel with nominal execution, an independent asynchronous supervisory process continuously monitors these sensory streams to assess execution safety and process consistency. When abnormal conditions are detected, such as fire ignition, overheating, ingredient burning, excessive seasoning, or pot overflow, the supervisor triggers corresponding intervention strategies. These interventions range from fine-grained corrective actions that adjust ongoing execution to emergency takeover that suspends or terminates the original cooking plan. Whenever possible, the module prioritizes graceful recovery by repairing the current process or adapting it toward an alternative dish that remains feasible under the updated state. The output of this module is a cooked dish ready for user consumption.

The post-cooking adaptation module supports long-term personalization by analyzing user feedback to inform future cooking decisions without affecting the current execution. Its input consists of multimodal post-meal feedback, including textual or spoken user comments and visual observations of leftover food, which are normalized into structured textual descriptions. These descriptions are analyzed to extract two types of information: execution- or recipe-related drawbacks that guide subsequent refinement of recipe preparation or online intervention strategies, and user-specific preferences such as selective eating habits or favored flavors. The output of this module is an updated set of historical dining records associated with the user, which are stored persistently and reused in later pre-cooking preparation stages to progressively improve recipe customization.

## IV. Pre-Cooking Preparation Module

This section details the procedures of the pre-cooking preparation module, presenting each step in sequence.

### *A. User Information Collection and Analysis*

The pre-cooking preparation module begins by collecting and analyzing user information together with the current kitchen context, producing a structured textual representation for subsequent recipe customization. Kitchen conditions are first characterized, including available cookware, heating equipment, ingredients, seasonings, and their quantities. User information is categorized into physiological and contextual factors. Physiological factors include age, health status, dietary restrictions, medical conditions, and nutritional considerations. Contextual factors encompass cultural and social influences, such as family background, ethnic traditions, regional culinary preferences, festival settings, special occasions, and other meal-related circumstances.

All inputs are categorized, labeled, and normalized into a unified textual format by lightweight intelligent agents, each responsible for a specific information category. At the current stage, most information is explicitly provided by the user through an interactive interface to ensure accuracy and transparency. In addition, historical dining records are incorporated as a persistent information source, summarizing prior dining behavior, satisfaction feedback, and leftover analysis. These records provide longitudinal context for understanding user preferences. The output of this step is a consolidated, interpretable textual description encoding cooking resources, user constraints, and contextual factors.

### *B. Canonical Recipe Generation*

Canonical recipe generation starts with dish name inference using a single-modal LLM driven by a carefully designed prompt template. The prompt explicitly constrains ingredient availability, seasoning scope, health safety, cultural appropriateness, and user intent alignment, ensuring that the generated dish name is feasible rather than merely plausible. To enforce output determinism, the prompt requires the model to return only a dish name within a fixed length and prohibits any auxiliary explanation.

To support domain-aware reasoning beyond generic language priors, this study constructed a specialized cooking knowledge dataset and used LoRA for efficient fine-tuning to obtain a language model with enhanced reasoning capabilities for the context of Chinese cooking. For detailed information, please refer to Section VII.

Once a dish name has been determined, a retrieval-augmented generation (RAG) pipeline is employed to construct the corresponding canonical recipe. The inferred dish name serves as a query to retrieve standardized references from a curated local knowledge base. When available, recipes defined by national or professional standards are prioritized; otherwise, widely recognized formulations are adopted. This process provides a consistent and authoritative foundation for subsequent recipe customization.

### *C. Customized Recipe Generation*

After the canonical recipe is determined, a customization procedure refines the canonical recipe to better match user-specific conditions. This procedure uses the aforementioned fine-tuned LLM again, but operates under a constrained setting in which the dish identity and canonical recipe are fixed. The model adjusts ingredients, seasonings, and cooking steps by incorporating personalized factors such as health constraints, dietary preferences, cultural considerations, and contextual requirements, while respecting resource availability. By framing customization as bounded modification of a canonical reference rather than open-ended generation, the process remains controllable and interpretable. The output is a structured personalized recipe with explicit quantities and a standardized format, documenting how the canonical recipe is adapted. Separating dish selection from recipe customization decomposes the reasoning process, avoids implicit end-to-end generation, and enables finer-grained supervision over personalization behavior.

### *D. Programmatic Recipe Generation*

This subsection describes how a customized recipe is transformed into a programmatic recipe through a two-stage reasoning process. The objective is to reformulate the customized recipe into a precisely structured intermediate material that bridges human-readable recipes and subsequent executable control code.

The transformation is carried out through two static AI agents, both implemented with fundamental LLMs. These LLMs are not fine-tuned for cooking knowledge, since domain understanding has already been addressed in earlier stages. Instead, they are employed for their general reasoning and structural transformation capabilities. The first LLM converts the customized recipe into an initial programmatic form that emphasizes stepwise control flow. Each cooking step is decomposed into a sequence of single-purpose operations, resulting in a finer and more explicit procedural structure than the original recipe description. At this stage, cooking actions are classified into two types, namely the instantaneous actions and continuous actions. Instantaneous actions correspond to one-off operations whose completion cannot be reliably sensed, such as ignition or ingredient addition, and are executed once before proceeding. Continuous actions correspond to operations that evolve over time, such as stirring or heating, and are expressed using explicit do–until constructs, where execution continues until terminal conditions are met.

For continuous actions, termination conditions are formulated using observable sensory criteria rather than implicit timing assumptions. Each condition may involve visual changes, temperature thresholds, sound intensity variations, or odor-related cues, depending on the nature of the operation. Specifically, for visual monitoring, the clarity of the ingredient state is quantified using

$$S_{focus^*}=Var(\nabla^2 I) \quad (1)$$

For acoustic monitoring, the system detects significant sound intensity shifts by comparing the moving average of the current window $W_{curr}$ against a past window $W_{past}$:

$$\Delta A=\frac{1}{|W_{curr}|}\int_{W_{curr}} A(t)\,dt-\frac{1}{|W_{past}|}\int_{W_{past}} A(t)dt>\theta \quad (2)$$

To ensure execution robustness, every continuous action is additionally associated with a maximum recommended duration derived from empirical cooking experience. This upper bound prevents indefinite execution in cases where sensory conditions are not satisfied, thereby guaranteeing progress and avoiding deadlock during online execution.

In the second stage, another fundamental LLM further normalizes and formalizes the initial programmatic recipe. All operations are rewritten using a predefined atomic action library, where each action follows a fixed verb–argument template and specifies explicit quantities, targets, and parameters. Ambiguous expressions are eliminated, and each step is constrained to contain exactly one minimal, independently executable action. This stage enforces strict consistency in wording, structure, and parameterization, ensuring that the programmatic recipe aligns with the capabilities of the cooking robot and sensing infrastructure while remaining human-readable and auditable.

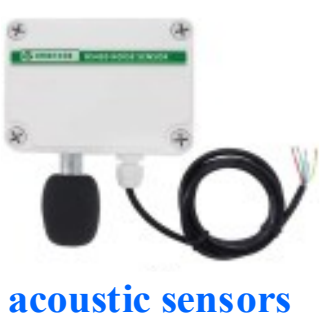
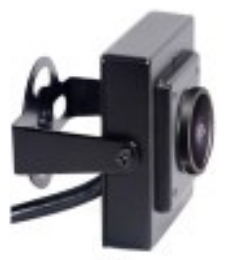
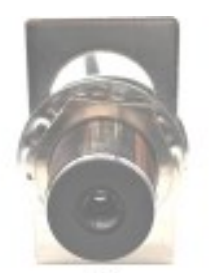
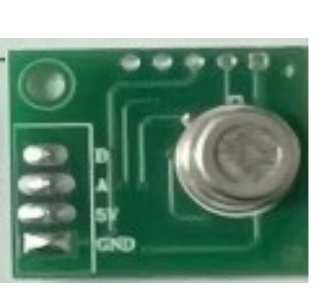

Fig. 2. Four types of sensors used in our automated cooking system.

As depicted in Fig. 2, the automated cooking system deploys four types of sensors. Visual sensors provide image-based observations of ingredient appearance, color changes, and spatial states within the cooking area. Thermal sensors, including infrared measurements, capture temperature distributions and overall heat levels, typically evaluated over spatial regions rather than single points.

To prevent localized scorching, we smooth the thermal matrix $R_t$ using an Exponential Moving Average (EMA):

$$M_t=\alpha R_t+(1-\alpha)M_{t-1}C_t \quad (3)$$

The thermal distribution is then analyzed by calculating the thermal centroid $C_t$:

$$x_c=\frac{\sum_{i,j} i\cdot w_{i,j}}{\sum_{i,j} w_{i,j}} \quad (4)$$

$$w_{i,j}=\max\left(0,T_{i,j}-T_{min}\right) \quad (5)$$

Furthermore, the dynamic state of the cooking process is determined by the trajectory magnitude $D_{move}$ derived from the covariance matrix $\Sigma$ of the centroid history:

$$D_{move}=\sqrt{\lambda_{max}(\Sigma)} \quad (6)$$

Acoustic sensors measure changes in sound intensity, which are used as auxiliary evidence and are generally evaluated based on relative variation rather than absolute values to account for background noise. Olfactory sensors are employed primarily for safety monitoring and late-stage supervision, such as detecting burning or characteristic odors during specific cooking phases.

## E. Executable Code Generation

This subsection describes how the programmatic recipe is compiled into a runnable Python source file that is directly executed in the online execution module. A fundamental large language model is employed here, as the task requires general code synthesis and control-flow construction rather than cooking-domain expertise. The input to this procedure is a structured programmatic recipe defined at the task level, where each step specifies a single cooking operation together with its progression logic, while low-level physical execution is delegated to a separate robotic agent responsible for continuous motion control and hardware interfacing.

The generated Python code follows a strictly sequential execution logic aligned with the programmatic recipe. Operations classified as instantaneous are rendered as explicit user-facing instructions and pause execution until user confirmation, whereas continuous operations are implemented as conditional loops that repeatedly evaluate termination criteria derived from sensor observations. These criteria are expressed in terms of visual states, temperature measurements, acoustic variations, and olfactory signals, and are combined through logical operators to determine step completion. To ensure robustness, each continuous step additionally enforces an upper time bound derived from the recommended duration specified in the programmatic recipe, allowing the execution to proceed even if sensor-based conditions are not met within a reasonable timeframe.

All sensing, condition-evaluation, and task-level actuation primitives are encapsulated as predefined interfaces whose implementations are fixed and provided to the LLM as contextual information. Consequently, the generated code is restricted to invoking these interfaces without modification, ensuring a consistent mapping between symbolic planning and physical execution.

### F. Complete Workflow of Pre-Cooking Preparation Module

This subsection describes the pre-cooking preparation workflow, which uses three internal mechanisms to improve reliability and transparency. After generating a recipe, the system verifies it against key constraints and reruns the process if needed. Once verified, the recipe is presented for user feedback and adjusted until approved. Finally, Python code is generated and validated, with iterative corrections made if errors occur.

## V. Online Execution Module

Online execution involves running the generated code as task-level actions dispatched to the robotic agent for step-by-step execution. Under normal conditions, this process runs deterministically. However, deviations may occur due to sensing uncertainty, actuation errors, or environmental changes. To resolve this issue, a real-time feedback and intervention mechanism operates asynchronously, monitoring perceptual data. If abnormalities are detected, it preempts the execution to initiate corrective or emergency actions. This section introduces the online supervisory system.

### A. Asynchronous Supervisory Architecture

The asynchronous supervisory process operates in parallel with the main execution thread, sharing the same multimodal sensory data stream while receiving feedback from the robotic execution layer. It produces structured intervention commands and parameter updates that are applied to the currently executing task through a unified control interface. To handle multiple abnormalities occurring simultaneously, explicit priority and mutual-exclusion rules are defined, allowing the supervisor to determine whether ongoing actions should be modified, suspended, overridden, or fully taken over. This design provides a consistent and deterministic mechanism for managing online interventions.

### B. Safety-Critical Intervention and Emergency Takeover

This subsection addresses safety-critical events that require immediate intervention and, if necessary, full takeover of the cooking process. The primary objective is to rapidly mitigate risk and prevent further state escalation in scenarios such as open flames, dry heating, severe burning indications, or violent overflow. Triggers are identified mainly through fast visual anomalies and abnormal thermal distributions, with olfactory cues and sound-intensity variations used as corroborating evidence. Once activated, the supervisory process prioritizes minimal decision latency and deterministic actions, including reducing or shutting off heat, suspending stirring or relocating cookware, and terminating the current operation to enter a safety mode. In extreme cases, the ongoing dish is abandoned altogether, reflecting the highest priority, strongest intervention strength, and maximal authority over the main execution flow.

### C. Recoverable Deviation Handling and Recipe Migration

This subsection separates online corrections into two branches, depending on whether the ongoing dish can still be preserved. If the deviation is recoverable, the goal is to diagnose the cause and repair the process with minimal, verifiable insertions while keeping the original dish objective unchanged. If preservation is no longer feasible, the goal shifts to recipe migration, where the process explicitly changes the dish objective and regenerates a feasible continuation that still satisfies the user's preference and health constraints.

For recoverable deviation handling, the supervisory process first performs cause inference by fusing multimodal sensory evidence with feedback from the robotic execution layer. In practice, this covers cases such as excessive seasoning, early-stage overflow, and uneven heating that signals a localized scorching tendency. For example, when a seasoning addition is over-executed and the liquid concentration or visual appearance deviates from the expected state, the supervisor inserts a bounded corrective step such as adding a measured amount of water. If the current salinity concentration $C_{curr}$ exceeds the target $C_{target}$, the required volume of water $\Delta V$ to be added is calculated by the dilution balance:

$$\Delta V = V_{curr}\left(\frac{C_{curr}}{C_{target}} - 1\right) \tag{7}$$

Followed by a short re-evaluation window, instead of rewriting the whole procedure.As another example, when infrared-derived heat maps indicate persistent hot spots within a spatial partition of the pan region, indicate persistent hot spots. To quantify the unevenness of the heating, we calculate the spatial variance $\sigma^2_{temp}$ of the temperature matrix $T$:

$$\sigma^2_{temp} = \frac{1}{H \times W} \sum_{i=1}^{H} \sum_{j=1}^{W} \left(T_{i,j} - \bar{T}\right)^2 \tag{8}$$

The supervisor injects a targeted redistribution action that focuses the robotic manipulator on the overheated region to disperse food and prevent further charring. In these recoverable cases, the inserted actions are intentionally small-step and pre-specified, so each correction can be validated quickly against updated sensory observations before the main procedure resumes.

For recipe migration, the supervisor activates a different branch only after determining that the dish objective cannot be restored without violating constraints or producing an unacceptable outcome. A representative trigger is a severe irreversible deviation, such as a level of over-seasoning that cannot be neutralized within safe and feasible bounds, or a state shift where the dish has effectively crossed into another preparation regime. In that situation, the supervisor proposes a new dish goal that is consistent with the current cooking state and does not violate user constraints, then regenerates the remaining steps accordingly so execution can continue without restarting from scratch. The migration branch therefore requires broader online reasoning than local repair, because it must reconcile what has already happened in the pan with what can still be achieved. Here, $D(\hat{s}, g)$ denotes the distance between the projected post-intervention cooking state $\hat{s}$ and the original dish goal state g, measured as a weighted sum of normalized deviations across key cooking variables including ingredient concentration, thermal distribution, and visual appearance. Formally, the system evaluates the reversibility by minimizing the distance metric $\mathcal{D}$ between the projected outcome state $S_{proj}$ and the original goal $S_{goal}$. The migration process is triggered only if the minimum achievable deviation exceeds the acceptable tolerance $\epsilon$:

$$\min_{\text{actions}} D\left(S_{proj}, S_{goal}\right) > \epsilon \tag{9}$$

while remaining compliant with user preferences and health constraints, and it updates the subsequent procedure in a synchronized manner so the running workflow transitions coherently rather than accumulating ad hoc patches.

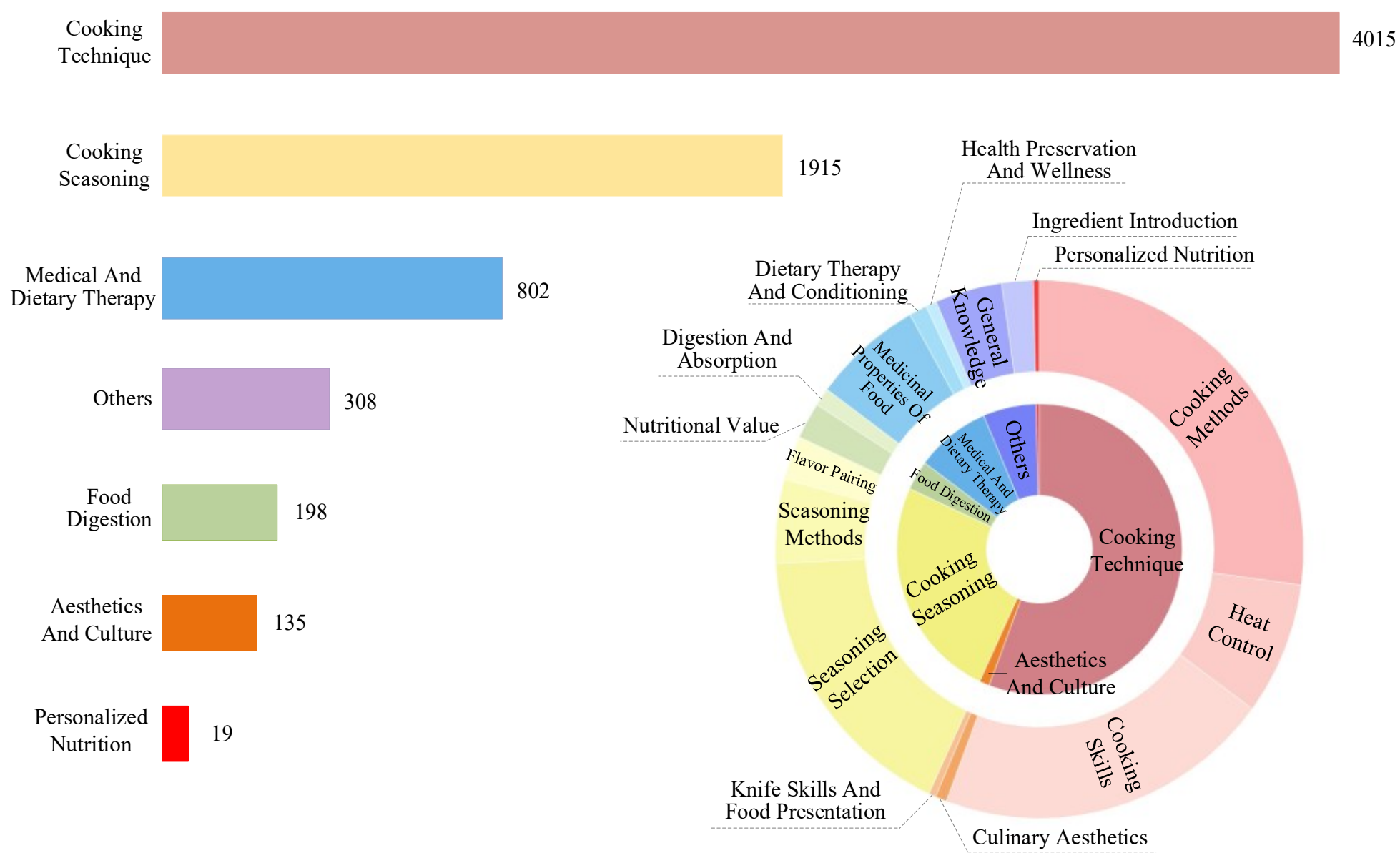

Fig. 3. Scale and structure of LLM training samples.

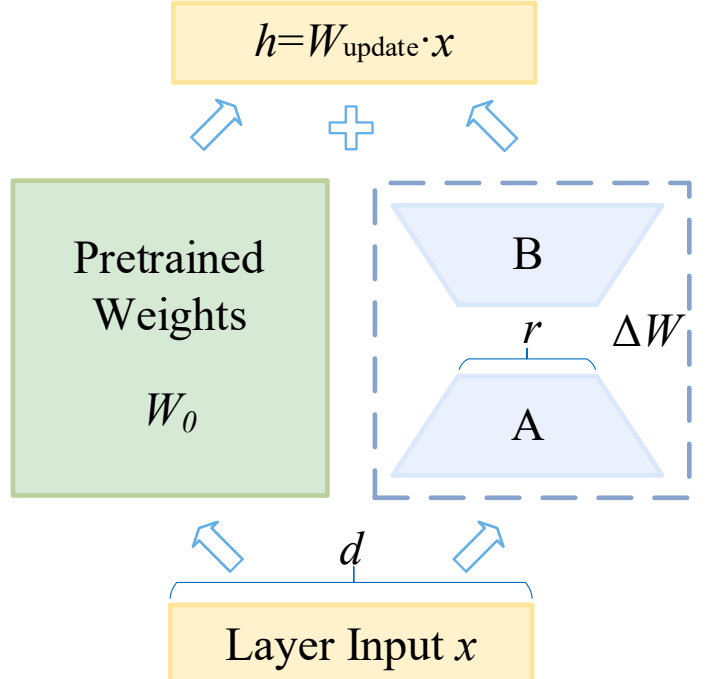

Fig. 4. Basic principle of lora fine-tuning.

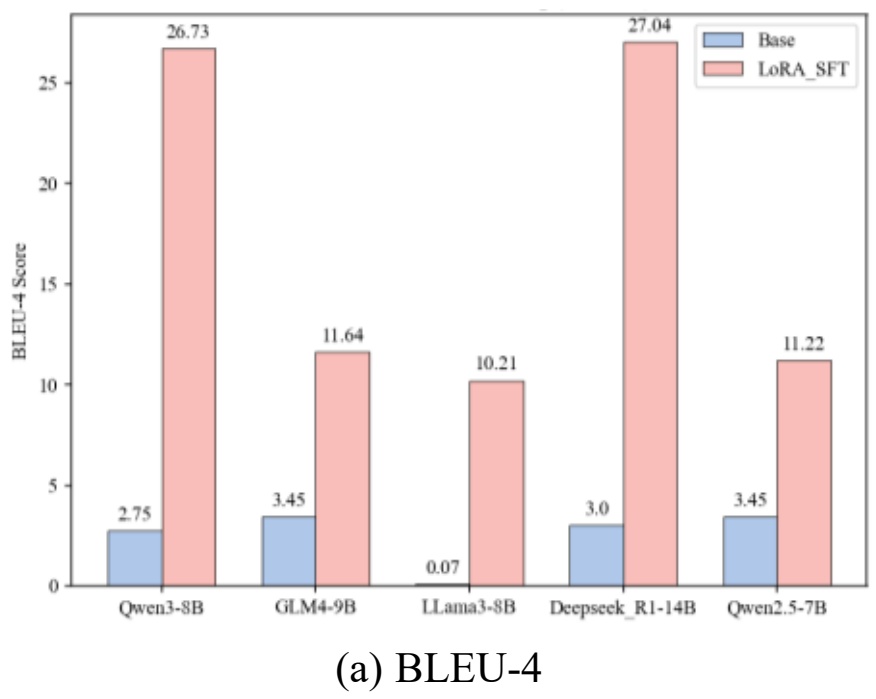

(a) BLEU-4

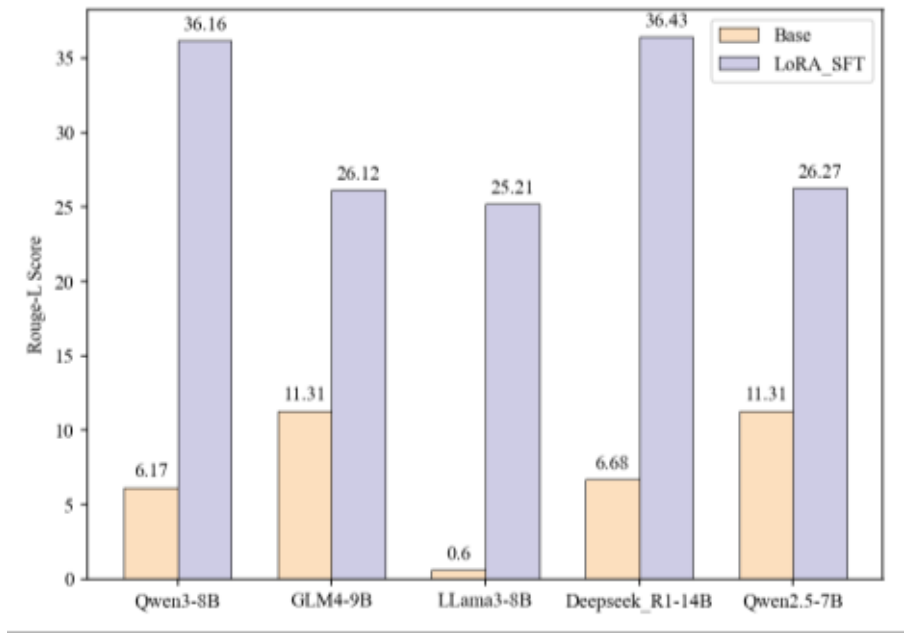

(b) Rouge-L

Fig. 5. Base vs. LoRA fine-tuning.

## VI. Post-Cooking Adaptation Module

This section addresses post-cooking adaptation, which captures user feedback after meal completion and converts it into structured knowledge for future use. The module does not affect the current dish. Instead, it consolidates multimodal observations and user responses into persistent records that guide subsequent recipe selection and customization.

One source of feedback is obtained through visual analysis of the finished dish. By examining the quantity and distribution of leftovers, the module distinguishes between portion-related effects and ingredient-specific consumption patterns. Uniform leftovers indicate excessive serving sizes, whereas repeated omission of particular ingredients suggests stable user preferences. The observations are interpreted together with historical dining records and, when necessary, supplemented by brief user interactions to distinguish short-term behaviors from persistent dietary tendencies.

The other source of feedback is explicit user input, collected through speech or text and normalized into textual form. This information is analyzed to separate dish-specific remarks from long-term preferences, such as recurring dislikes or favored flavors. The resulting conclusions are stored in the historical dining records at multiple levels, linking concrete outcomes to individual dishes while also updating user-level attributes that persist across meals. This layered memory supports progressively improved personalization while preserving transparent and predictable cooking behavior.

## VII. LoRA-Based SFT Fine-Tuning

This section provides an introduction to the construction process of the specialized cooking knowledge dataset and the content of the model fine-tuned efficiently using LoRA.

### *A. Dataset*

This dataset is designed to extract and integrate knowledge from Chinese culinary arts through an automated process.

Using advanced large language models, valuable information is extracted from literature and converted into a structured Q&A format with custom system prompts. This approach improves the efficiency and accuracy of knowledge acquisition, supporting personalized dietary recommendations, health management, and intelligent Q&A systems.

#### 1) *Automated Dataset Construction*

The dataset is built from 1,120 cooking-related articles retrieved from CNKI. These articles cover a broad range of topics, including cooking techniques, culinary aesthetics, nutritional value of ingredients, and theories of Chinese medicinal diets. We extract text from these documents via the model-driven pipeline and convert it into structured data.

To achieve efficient batch processing of the literature, this study developed a file processing agent. This agent invokes PDF parsing tools and integrates natural language processing techniques to filter and extract relevant content. To evaluate the overall quality of the automatically constructed dataset, this study measures the semantic consistency between questions and answers. Specifically, we define the dataset quality metric, denoted as $Q_{quality}$, as the mean semantic similarity across all instruction-response pairs:

$$Q_{quality}=\frac{1}{n}\sum_{i=1}^{n} Similarity(Q_i,A_i) \tag{10}$$

$$Similarity(Q_i,A_i)=\frac{e(Q_i)\cdot e(A_i)}{\|e(Q_i)\|\cdot\|e(A_i)\|} \tag{11}$$

Herein, $Q_i$ and $A_i$ represent the $i$-th question and its corresponding answer, and $n$ is the total number of question-answer pairs in the dataset. The similarity function measures how well the question and answer match in semantic space, and can in principle be computed using vector representations from a pre-trained language model e(·), such as bge-m3 or text-embedding-3-small. This metric provides a theoretical measure of the consistency and reliability of the automatically generated question-answer pairs; empirical computation of this metric is left for future work.

#### 2) *Cooking Knowledge Instruction Dataset*

This dataset focuses on Chinese cooking and covers seven major categories, including cooking techniques, seasoning and ingredient pairing, medical and dietary therapy, food digestion and metabolism, culinary aesthetics and culture, personalized nutrition, and other practical cooking knowledge. The dataset encompasses topics such as ingredient preparation, heat control, flavor design, medicinal food applications, health-oriented dietary recommendations, visual and sensory characteristics of Chinese cuisine, and practical skills related to cooking tools and ingredient handling.

#### 3) *Dataset Size*

This dataset contains a total of 7,392 entries. In terms of distribution, the largest proportion is in cooking techniques, accounting for 54.32%, followed by seasoning and flavoring at 25.91%, and medicinal food at 10.85%. The remaining sections, including others, food digestion and metabolism, aesthetics and culture, and personalized diets, each have smaller proportions. The content of each section provides comprehensive data support for the field of Chinese cooking and personalized healthy eating. The detailed distribution is shown in Fig. 3.

TABLE I
COMPARISON OF BASE AND LORA FINE-TUNED MODELS

| Model | | BLEU-4 | Rouge-L |
|---|---|---|---|
| Qwen3-8B | Base | 2.75 | 6.17 |
| | LoRA_SFT | 26.73 | 36.16 |
| GLM4-9B | Base | 3.45 | 11.31 |
| | LoRA_SFT | 11.64 | 26.12 |
| Llama3-8B | Base | 0.07 | 0.60 |
| | LoRA_SFT | 10.21 | 25.21 |
| Deepseek-14B | Base | 3.00 | 6.68 |
| | LoRA_SFT | 27.04 | 36.43 |
| Qwen2.5-7B | Base | 3.45 | 11.31 |
| | LoRA_SFT | 11.22 | 26.27 |

### B. *Domain-Specific Model Fine-Tuning*

We applied supervised fine-tuning to multiple pre-trained models using the LoRA (Low-Rank Adaptation) method, and utilized them for text generation tasks in the domain of Chinese cooking. The core idea of the LoRA method is to introduce low-rank decomposition matrices into the Transformer layers while freezing the pre-trained model weights, thereby significantly reducing the number of parameters that need to be adjusted during fine-tuning [25].

This strategy not only effectively reduces computational and storage overhead but also provides an efficient solution for the model to quickly adapt to new tasks and domain shifts. Specifically, we fine-tuned the language layers of the model while freezing the weights of the non-language layers. Through this process, the model is able to better understand and learn the textual knowledge related to Chinese cooking, and generate personalized recipe recommendations, thus improving both task execution efficiency and performance. The basic principle is illustrated in Fig. 4, where the LoRA method achieves efficient fine-tuning by introducing low-rank bias terms $\Delta W$ into the cooking model's weight matrix. The fine-tuned model parameter matrix $W_{update}$ can be expressed as

$$W_{\text{update}}=W_0+\Delta W \tag{12}$$

where $W_0 \in \mathbb{R}^{d\times k}$ is the pre-trained parameter matrix before fine-tuning, with $d$ and $k$ representing the number of rows and columns of the original matrix, respectively. By adding the pre-trained parameter matrix $W_0$ and the low-rank bias term $\Delta W$, the fine-tuned parameter matrix is obtained. The bias term $\Delta W$ is further expressed as

$$\Delta W=B\cdot A \tag{13}$$

where $B\in\mathbb{R}^{d\times r}$ and $A\in\mathbb{R}^{r\times k}$ are two low-rank matrices, and $r\ll \min(d,k)$ represents the intrinsic rank of $\Delta W$. During training, only the parameters of $A$ and $B$ are updated, while $W_0$ remains frozen. This approach enables efficient supervised fine-tuning of the model while significantly reducing computational and storage costs. At this point, the model's offline fine-tuning process is complete.

## VIII. EXPERIMENTAL RESULTS AND DISCUSSIONS

This section validates the proposed agentic cooking framework through three progressively advancing experiments. First, the performance of the fine-tuned domain-specific model is evaluated in personalized recipe generation. Second, the effectiveness of the proposed recipe-to-code generation pipeline is assessed through automated code generation experiments. Finally, physical experiments are conducted to verify the feasibility, reliability, and practical applicability of the overall framework in real-world environments.

TABLE II
COMPARISON OF TWO CODE GENERATION METHODS ON DISH CONTROL CODE METRICS

| Dishes | End-to-End | | | Step-by-Step | | |
|---|---|---|---|---|---|---|
| | Executability | Functional Correctness | Structural Integrity | Executability | Functional Correctness | Structural Integrity |
| Sweet and Sour Chinese Cabbage | √ | 100 | 100 | √ | 100 | 100 |
| Sautéed Beef with Green Peppers | × | 50 | 100 | √ | 100 | 100 |
| Sautéed Lamb with Scallions | × | 50 | 75 | √ | 100 | 100 |
| Kung Pao Chicken | × | 50 | 100 | √ | 100 | 100 |
| Sautéed Shelled Shrimps | × | 50 | 100 | √ | 100 | 100 |
| Mapo Tofu | × | 50 | 75 | √ | 100 | 100 |
| Scrambled Eggs with Tomatoes | × | 50 | 100 | √ | 100 | 100 |
| Braised Eggplant in Brown Sauce | × | 75 | 100 | √ | 100 | 100 |
| Hot & Sour Shredded Potatoes | × | 75 | 75 | √ | 100 | 100 |
| Twice-Cooked Pork | √ | 100 | 100 | √ | 100 | 100 |
| Sautéed Three Fresh Vegetables | × | 75 | 75 | √ | 100 | 100 |
| Sautéed Black Fungus with Celery | × | 50 | 100 | √ | 100 | 100 |
| Dry Pot Cabbage | √ | 100 | 100 | √ | 100 | 100 |
| Celery & Dried Tofu Sautéed Pork | × | 50 | 75 | √ | 100 | 100 |
| Braised Pork Ribs in Brown Sauce | × | 50 | 75 | √ | 100 | 100 |
| Sautéed Ham with Garlic Sprouts | √ | 100 | 100 | √ | 100 | 100 |
| Braised Green Beans with Potatoes | × | 50 | 100 | √ | 100 | 100 |
| Sautéed Broccoli with Garlic | × | 75 | 100 | √ | 100 | 100 |
| Sautéed Leaf Lettuce | × | 50 | 100 | √ | 100 | 100 |
| Steamed Perch | × | 50 | 100 | √ | 100 | 100 |
| Score | 20% | 65 | 92.5 | 100% | 100 | 100 |

### *A. Benchmarking of Fine-Tuning in Cooking LLM*

Five pre-trained language models of different sizes were fine-tuned using LoRA on the cooking instruction dataset. The models were evaluated for their adaptability and knowledge enhancement in the cooking domain using BLEU-4 and ROUGE-L metrics. BLEU-4 measures the n-gram overlap between generated and reference texts, while ROUGE-L evaluates the longest common subsequence match. Higher scores indicate greater similarity between generated and reference texts, with a maximum score of 100.

To meet the high load requirements of parallel computing and deep learning inference, this experiment used a server equipped with two NVIDIA A100-SXM4 80GB GPUs as the core computing node, running the Ubuntu 20.04 operating system, and with third-party computing libraries such as Python and TensorFlow installed.

The experimental results are shown in the Fig. 5 and Table I below. The model fine-tuned with LoRA significantly outperforms the baseline model in both BLEU-4 and ROUGE-L metrics. Specifically, the fine-tuned Qwen3-8B achieved a BLEU-4 score of 26.73, a 23.98% improvement over the baseline model (2.75), and a ROUGE-L score of 36.16, a 23.51% improvement over the baseline model (6.17). Similarly, GLM4-9B also showed significant improvements in both metrics, with a BLEU-4 increase of 8.19% and a ROUGE-L increase of 14.81%. For Llama3-8B, although the baseline score was lower, the fine-tuned model showed a 10.14% increase in BLEU-4 and a 24.61% increase in ROUGE-L. On Deepsseek_R1_Distill_qwen_14B, the BLEU-4 and ROUGE-L improvements reached 24.04% and 29.75%, respectively, demonstrating a substantial performance boost after fine-tuning. For Qwen2.5-7B, BLEU-4 improved by 7.77%, and ROUGE-L improved by 14.96%. These results indicate that the LoRA fine-tuning strategy significantly enhances the performance of each model on cooking domain tasks, particularly in generating personalized Chinese recipes, fully validating the effectiveness of this method.

### *B. Comparisons Among Automated Code Generation Methods*

This experiment compares the performance of the end-to-end one-shot generation method with the staged automated generation method on a test set containing 20 dish-control code mappings. The end-to-end method generates the control code in a single step using a single prompt, without requiring intermediate interactions. The accuracy and stability of the generated code are evaluated using metrics such as executability, functional correctness, and structural integrity. Executability refers to whether the code runs correctly without errors; functional correctness measures the completeness and accuracy of core structures; structural integrity focuses on the compliance of the code's syntax and module dependencies. All metrics are scored out of 100. The experimental results are shown in Table II.

The results show that in the 20-dish test set, the end-to-end generation method had an executability rate of only 20%, with an average functional correctness score of 65 and an average structural integrity score of 92.5. Many dishes experienced issues such as runtime failures and functional errors. In contrast, the staged generation method achieved a perfect score of 100% across all three metrics, with stable execution, error-free functionality, and compliant structure for all dish codes. This demonstrates that the staged strategy effectively addresses the semantic overload issue of the end-to-end method, significantly improving the accuracy and stability of dish control code generation.

### C. *Physical Experiment Validation*

To validate the feasibility of the physical experiment, this study was conducted in a real kitchen environment, equipped with a gas stove (with a heat range of 1.2kW for low heat, 2.8kW for medium heat, and 4.2kW for high heat), a standard wok (32 cm in diameter, 3 mm thick), and supporting kitchen tools. The experiment also employed multimodal sensors for process monitoring, including visual (S-YUE SY011HD), infrared thermal imaging (MELEXIS MLX90640), audio (SOONSALLXM8165B), and taste (ZP16) sensors. The overall setup is shown in Fig. 6.

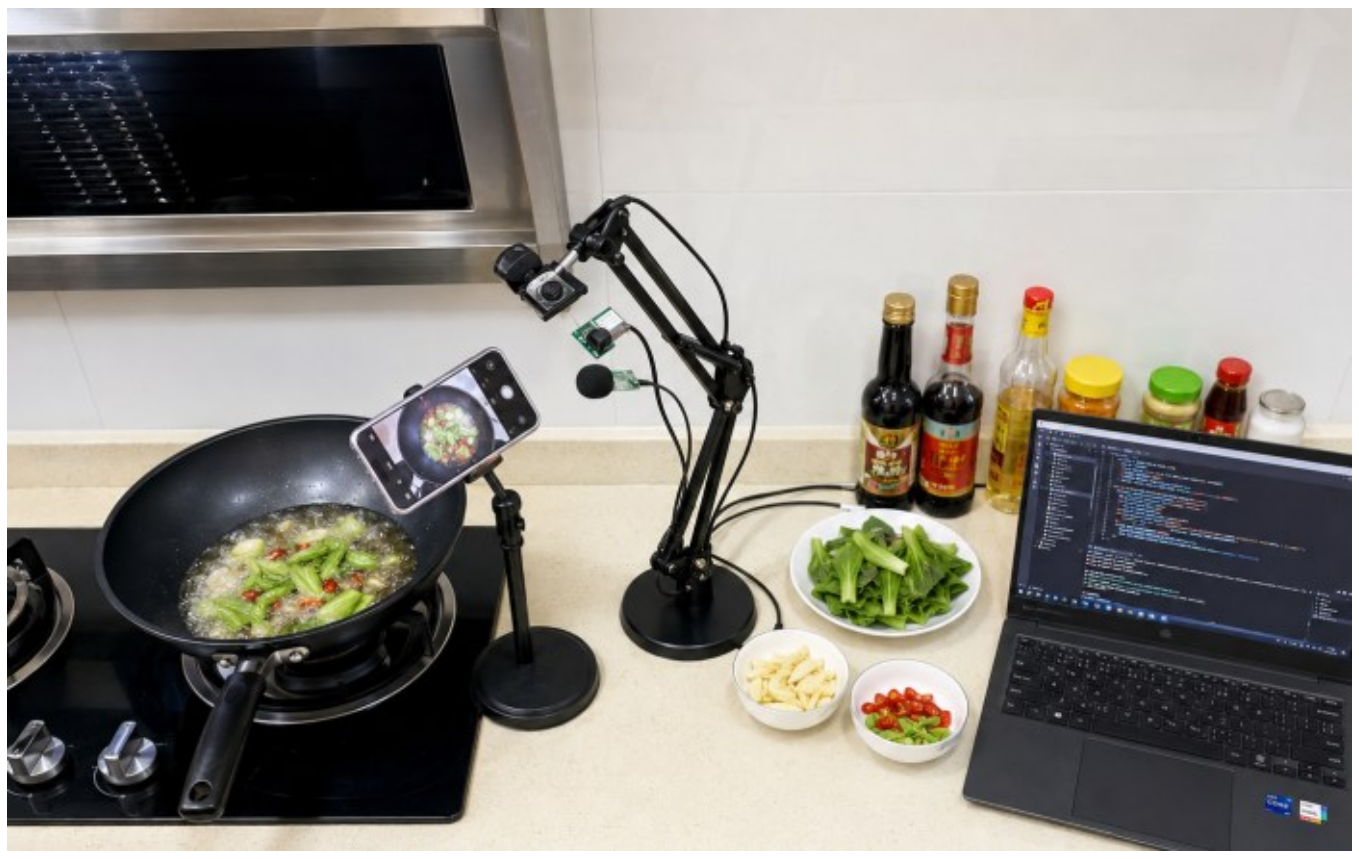

Fig. 6. Experimental setup of automated cooking platform.

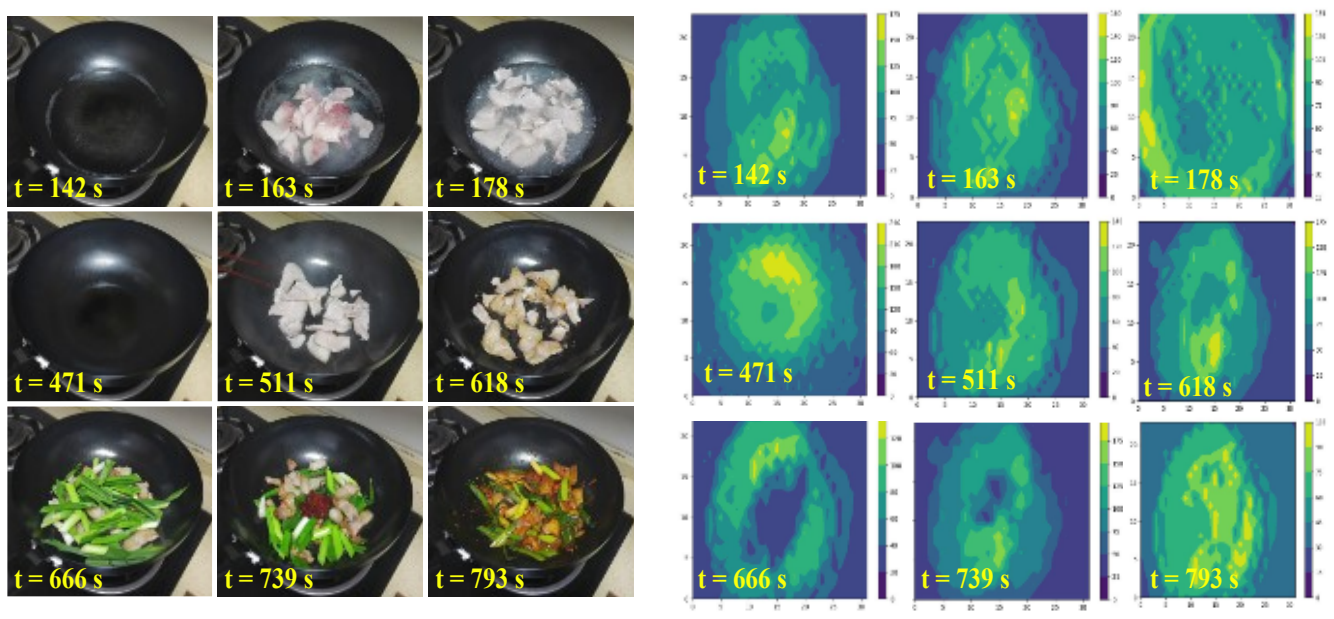

Fig. 7. Visual and thermal changes in Salted Pork.

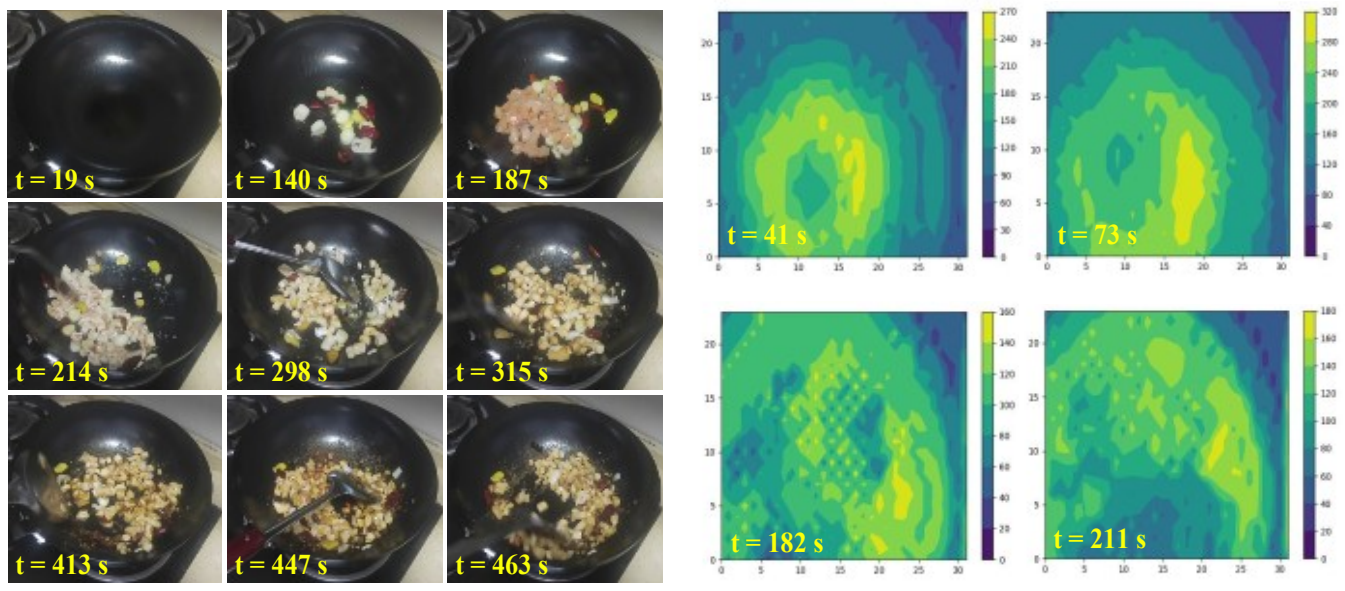

Fig. 8. Visual and thermal changes in Kung Pao Chicken.

We selected two typical Chinese dishes, Salted Pork and Kung Pao Chicken, as the validation targets. Salted Pork involves cooking steps such as heating the pan, stir-frying, and seasoning integration, while Kung Pao Chicken includes multi-stage operations like deep frying, stir-frying, and mixing, involving complex physical processes such as high-temperature phase transitions and heat control. These dishes are suitable for validating the ability to provide precise guidance for continuous cooking actions. Figs. 7 and 8 shows the changes during the cooking process of the two dishes.

To test abnormal scenario handling, we simulated a fire on the oil surface. The system, using visual and temperature sensors, triggered an emergency intervention within 120ms, turning off the heat source and covering the fire, effectively preventing escalation.

The experimental results show the system successfully completes the process from user requirements to executable code, with 100% successful execution. The instructions clearly specify details like ingredient quantities, heat levels, and durations. These results demonstrate that the multi-agent decomposition framework effectively converts recipes into practical cooking processes that meet real-world needs.

## IX. Conclusions

This paper proposes an integrated agentic framework that embeds Large Language Models into flow controls to address the critical challenges of personalization, transparency, and online adaptivity in automated cooking. By establishing a closed-loop architecture comprising pre-cooking preparation, online execution, and post-cooking adaptation, the system successfully decomposes complex user requirements into executable, verifiable control codes via a multi-agent approach, effectively overcoming the reliability issues inherent in black-box end-to-end generation. The integration of an asynchronous supervisory mechanism with heterogeneous multimodal sensors further ensures execution safety and enables real-time recovery from deviations through dynamic intervention strategies. Experimental results confirm that the LoRA-fine-tuned model significantly enhances domain-specific reasoning, the staged code generation strategy achieves 100% executability, and the physical platform capably performs complex Chinese cooking tasks with robust handling of safety anomalies, thereby providing a trustworthy and scalable solution for intelligent culinary automation. It should be noted that the current system still requires user confirmation for certain instantaneous operations whose completion cannot yet be reliably verified through existing sensors. This represents a deliberate trade-off between execution safety and full automation: manual confirmation ensures that safety-critical one-off actions are not executed under uncertain conditions, while inevitably introducing human intervention into an otherwise automated pipeline. In future work, advances in multimodal perception are expected to enable autonomous verification of these operations, progressively reducing the need for manual confirmation and bringing the system closer to complete end-to-end automation.

## Acknowledgment

The authors would like to acknowledge the support of the Qingdao Academy of Intelligent Industries (QAII) and its affiliated institutes, including the Institute of Smart Healthcare, for their support of a series of research initiatives since 2014, such as Cuisine 5.0 and the Yiyin Large Model, with related experiments conducted in Qingdao, Beijing, Weihai, and other locations. These initiatives have aimed to advance the transformation of livelihood-related industries, including healthcare, clothing, food, housing, and transportation, through AI and intelligent technologies for the benefit of humanity. The authors also acknowledge the Parallel Chef project, formally launched by QAII in 2021, and the Yiyin MeTaurant research program, established since 2022 and led by Prof. Bai Li, Yonglin Tian, and Fei-Yue Wang, which focuses on the innovative application of emerging AI technologies, including the metaverse, large language models, AI agents, retrieval-augmented generation, agentic intelligence, and autonomous intelligence, in livelihood-related domains. This article presents part of the preliminary outcomes of these projects. The authors extend their sincere gratitude to all researchers and engineers who have contributed to the discussions, research, and experiments over the past decade.

## References

[1] F. Seyitoğlu, F. Fusté-Forné, S. Yiğit, and S. Engin, "Robot chefs: The impacts, compatibility and suitability," *British Food Journal*, vol. 127, no. 1, pp. 307–323, 2025.

[2] J. D. Blutinger, C. C. Cooper, S. Karthik, A. Tsai, N. Samarelli, E. Storvick, et al., "The future of software-controlled cooking," *Nature Science of Food*, vol. 7, no. 1, art. 6, 2023.

[3] D. Pereira, A. Bozzato, P. Dario, and G. Ciuti, "Towards foodservice robotics: A taxonomy of actions of foodservice workers and a critical review of supportive technology," *IEEE Transactions on Automation Science and Engineering*, vol. 19, no. 3, pp. 1820–1858, 2022.

[4] C. L. Lee, and H. S. Kwak, "Effect of cooking and food serving robot design images and information on consumer liking, willingness to try food, and emotional responses," *Food Research International*, art. 116626, 2025.

[5] G. Sochacki, X. Zhang, A. Abdulali, and F. Iida, "Towards practical robotic chef: Review of relevant work and future challenges," *Journal of Field Robotics*, vol. 41, no. 5, pp. 1596–1616, 2024.

[6] N. K. Mondal, M. G. Rashed, D. Das, and A. Z. M. T. Islam, "RAJUNI: Design and implementation of an autonomous cooking robot," in *Proc. IEEE 27th Int. Conf. Computer and Information Technology (ICCIT)*, Dec. 2024, pp. 1880–1885.

[7] W. X. Yan, Z. Fu, Y. H. Liu, Y. Z. Zhao, X. Y. Zhou, J. H. Tang, and X. Y. Liu, "A novel automatic cooking robot for Chinese dishes," *Robotica*, vol. 25, no. 4, pp. 445–450, 2007.

[8] W. T. Ma, W. X. Yan, Z. Fu, and Y. Z. Zhao, "A Chinese cooking robot for elderly and disabled people," *Robotica*, vol. 29, no. 6, pp. 843–852, 2011.

[9] D. Noh, H. Nam, K. Gillespie, Y. Liu, and D. Hong, "Yummy operations robot initiative: An autonomous cooking system utilizing a modular robotic kitchen and dual-arm proprioceptive manipulator," *IEEE Robotics & Automation Magazine*, doi: 10.1109/MRA.2025.3615353.

[10] Y. Sugiura, D. Sakamoto, A. Withana, M. Inami, and T. Igarashi, "Cooking with robots: Designing a household system working in open environments," in *Proc. SIGCHI Conf. Human Factors in Computing Systems*, Apr. 2010, pp. 2427–2430.

[11] K. Junge, J. Hughes, T. G. Thuruthel, and F. Iida, "Improving robotic cooking using batch Bayesian optimization," *IEEE Robotics and Automation Letters*, vol. 5, no. 2, pp. 760–765, 2020.

[12] G. Sochacki, A. Abdulali, N. K. Hosseini, and F. Iida, "Recognition of human chef's intentions for incremental learning of cookbook by robotic salad chef," *IEEE Access*, vol. 11, pp. 57006–57020, 2023.

[13] G. Sochacki, J. Hughes, S. Hauser, and F. Iida, "Closed-loop robotic cooking of scrambled eggs with a salinity-based taste sensor," in *Proc. IEEE/RSJ Int. Conf. Intelligent Robots and Systems (IROS)*, Sep. 2021, pp. 594–600.

[14] J. Shi, A. Abdulali, G. Sochacki, and F. Iida, "Closed-loop robotic cooking of soups with multi-modal taste feedback," in *Proc. Annu. Conf. Towards Autonomous Robotic Systems*, Sep. 2023, pp. 51–62.

[15] D. Driess, F. Xia, M. S. Sajjadi, C. Lynch, A. Chowdhery, B. Ichter, et al., "PaLM-E: An embodied multimodal language model," in *Proc. 40th Int. Conf. Machine Learning (ICML)*, Jul. 2023, pp. 8469–8488.

[16] B. Zitkovich, T. Yu, S. Xu, P. Xu, T. Xiao, F. Xia, et al., "RT-2: Vision-language-action models transfer web knowledge to robotic control," in *Proc. Conf. Robot Learning (CoRL)*, Dec. 2023, pp. 2165–2183.

[17] F. Gravot, A. Haneda, K. Okada, and M. Inaba, "Cooking for humanoid robot, a task that needs symbolic and geometric reasonings," in *Proc. IEEE Int. Conf. Robotics and Automation (ICRA)*, May 2006, pp. 462–467.

[18] M. Schmitz, F. Menz, R. Grunau, N. Mandischer, M. Hüsing, and B. Corves, "Robot cooking—Transferring observations into a planning language: An automated approach in the field of cooking," *Eng*, vol. 4, no. 4, pp. 2514–2524, 2023.

[19] M. Inagawa, T. Takei, and E. Imanishi, "Analysis of cooking recipes written in Japanese and motion planning for cooking robot," *Robomech Journal*, vol. 8, no. 1, art. 17, 2021.

[20] S. H. Vemprala, R. Bonatti, A. Bucker, and A. Kapoor, "ChatGPT for robotics: Design principles and model abilities," *IEEE Access*, vol. 12, pp. 55682–55696, 2024.

[21] J. Liang, W. Huang, F. Xia, et al., "Code as policies: Language model programs for embodied control," in *Proc. IEEE Int. Conf. Robotics and Automation (ICRA)*, London, United Kingdom, 2023, pp. 9493–9500.

[22] T. Yoneda, J. Fang, P. Li, et al., "Statler: State-maintaining language models for embodied reasoning," in *Proc. IEEE Int. Conf. Robotics and Automation (ICRA)*, Yokohama, Japan, 2024, pp. 15083–15091.

[23] A. Mavrogiannis, C. Mavrogiannis, and Y. Aloimonos, "Cook2LTL: Translating cooking recipes to LTL formulae using large language models," in *Proc. IEEE Int. Conf. Robotics and Automation (ICRA)*, Yokohama, Japan, 2024, pp. 17679–17686.

[24] M. Verghese and C. Atkeson, "Skills made to order: Efficient acquisition of robot cooking skills guided by multiple forms of internet data," in *Proc. IEEE Int. Conf. Robotics and Automation (ICRA)*, Atlanta, USA, 2025, pp. 11965–11971.

[25] Hu E J, Shen Y, Wallis P, et al. Lora: Low-rank adaptation of large language models[J]. ICLR, 2022, 1(2): 3.